\documentclass[a4paper, 11pt]{article}
\usepackage{graphicx} 
\usepackage{hyperref}
\usepackage[estonian]{babel}
\usepackage{fullpage}

\title{{\bf Eestikeelse teksti automaatkorrektuur}:\\ projekti EKTB25 lõpparuanne}
\author{Agnes Luhtaru, Martin Vainikko, Krista Liin, Mark Fišel (Tartu Ülikool)\\
Kais Allkivi-Metsoja, Jaagup Kippar, Pille Eslon (Tallinna Ülikool)}
\date{15. veebruar 2024}

\begin{document}

\newcommand{\fsc}{F$_{0.5}$}

\maketitle

Projekti rahastas aastail 2021-2023 Eesti keeletehnoloogia riiklik programm. Selle põhieesmärk oli arendada eesti keele jaoks õigekirja- ja grammatikakorrektuuri vahendeid. Peamine väljakutse oli väga väike hulk olemasolevaid veaparanduse andmeid, mida on tarvis sellise arendustöö jaoks. Selle leevendamiseks

\begin{itemize}
\item täiendasime parandusandmeid mudelite treenimiseks ja testimiseks,
\item katsetasime siirdeõpet ehk teiste ülesannete jaoks treenitud masinõppe\-mudelite ümberõpetamist, et mitte sõltuda ainult parandusandmetest,
\item võrdlesime arendatud meetodit ja mudelit alternatiividega, sh suurte keelemudelitega
\end{itemize}

Samuti arendasime automaatset hindamist, mille abil saab arvutada paranduste täpsust ja saagist veakategooriate kaupa, et saaks detailselt võrrelda erinevate meetodite efektiivsust.

Projekti kestel on toimunud suurte keelemudelite läbimurre. On loodud GPT4, kommertskeelemudel eestikeelse toega. Võtsime mudeli olemasolu arvesse plaanide kohendamisel ning esitame aruandes võrdluse GPT4 võimega parandada eestikeelset teksti.

Lõpptulemused näitavad, et meie arendatud lähenemine annab paremaid skoore kui GPT4 ning tulemus on kasutatav, kuid pole siiski täiesti usaldusväärne. Aruanne sisaldab ka ideid selle kohta, kuidas GPT4 ja teisi suuri keelemudeleid edaspidi rakendada saaks, seejuures keskendudes vabavarana kasutatavatele lahendustele.

Kõik käesoleva projekti tulemid on avaandmed/vabavara; litsentsid võimalda\-vad nende kasutamist mh ka kommertseesmärkidel:

\begin{itemize}
    \item Grammatika- ja õigekirjakontrolli kood\footnote{\url{https://koodivaramu.eesti.ee/tartunlp/corrector}}
    \item Grammatikakontrolli\footnote{\url{https://huggingface.co/tartuNLP/en-et-de-cs-nelb}} ja õigekirjakontrolli mudelid: \footnote{\url{https://huggingface.co/Jaagup}}
    \item Treeningandmed\footnote{\url{https://github.com/TartuNLP/estgec/tree/main/Tartu_L2_corpus}}, arendus- ja testandmed (K2)\footnote{\url{https://github.com/tlu-dt-nlp/EstGEC-L2-Corpus}} , testandmed (K1)\footnote{\url{https://github.com/TartuNLP/estgec/tree/main/Tartu_L1_corpus}}
\end{itemize}

Aruanne koosneb neljast peatükist. \ref{sct:data}. peatükis anname ülevaate treening- ja testandmetest, \ref{sct:meth}. peatükis kirjeldame kasutatud metoodikat, \ref{sct:eval}. peatükis esitame korrektuurivahendite hindamise tulemused ning \ref{sct:fut}. peatükis toome välja võimalikud suunad eestikeelse teksti automaatkorrektuuri edasiarenduseks.

\newpage

\section{Andmed}
\label{sct:data}

\subsection{Veamärgendusega test- ja arenduskorpus}

Eesti keele grammatikakontrollija arendamiseks ja testimiseks hakkasime olemasolevate tekstikogude põhjal koostama ühtse veamärgendusega korpust, mis peaks sisaldama suhteliselt võrdsel määral eesti keele kui teise keele (K2) õppijate ja eesti keelt emakeelena (K1) kõnelejate toimetamata kirjutisi. Korpusesse valisime esialgu üle 7000 lause, mis on mahult võrreldav grammatikakorrektorite võistluse BEA-2019 arendus- ja testandmestikuga (Bryant et al., 2019). Praeguseks oleme märgendanud ligi 5000 lauset. Vigade asukoha, liigi ja paranduse tähistamiseks lausete kaupa oleme kasutanud M2-märgendusformaati (Dahlmeier \& Ng, 2012). Vealiigituse aluseks võtsime ERRANT-i klassifikatsiooni (Bryant et al., 2017), mida kohandasime eesti keele jaoks.

Üldisem vigade liigitus tugineb sellele, kas sõna või kirjavahemärk on liigne, puudub või vajab asendamist muu sõnakuju või kirjavahemärgiga. Sõnaasendused jagunevad omakorda täheortograafia-, algustähe-, kokku-lahkukirjutuse, käänd- ja tegusõna vormivaliku, sõnavaliku- ja sõnajärjevigadeks (asendatakse mitmesõnaline üksus). Juhtudel, kus samas sõnas on mitu viga, kasutame liitveamärgendeid. Põhimärgendeid on 12, liitmärgendeid lisaks 18. Näiteks võib vormi- või sõnavalikuviga langeda kokku eksimusega tähe\-orto\-graa\-fias või kokku- ja lahkukirjutuses. Kui ebaloomuliku sõnajärjega lauseosa hõlmab ka mõnda sõnatasandi viga, siis märgime need omaette vigadena, et nende parandamise üle saaks eraldi arvestust pidada. Ühele lausele lisame kuni kolm märgendusversiooni. Selleks oleme teisendanud eelneva veamärgenduse ja seda täiendanud.

K2 materjal (263 teksti, 3790 lauset) pärineb Eesti vahekeele korpusest (EVKK) ja jaotub nelja keeleoskustaseme (A2, B1, B2 ja C1) vahel. Need tekstid olid varem märgendatud CoNLL-U formaadis, kus morfosüntaktilise info kõrval oli välja toodud veainfo koos parandustega – üks märgendusversioon lause kohta. Selles formaadis testandmeid on kasutatud eesti keele õigekirjakontrolli vahendite võrdlemiseks, kuid grammatikakontrolli tulemusi on kasulik hinnata mitmekülgsema märgenduse alusel. Testandmestiku moodustavad 2029 lauset. Materjalist, mida plaanime rakendada edasises arendustöös, on märgendatud 1363 lauset. 

K1 tekstimaterjal (81 teksti, 3546 lauset) on võetud eesti keele õppija korpuse EMMA eesti emakeelega keskkooliõpilaste eksamikirjutiste alamhulgast, kus on märgendatud õigekirja-, kirjavahemärgi-, stiili- ja faktivead. Kuna EMMA märgenduses on vealiigitus üldisem ning seal pole välja pakutud parandusi, siis täpsustasime vealiigi oma märgendusskeemis ning lisasime võimaliku paranduse. Sealjuures on ühe vea kohta eri märgendajatelt eri parandusversioonid. Osad algselt märgendatud vead jäid vaatluse alt välja, nt faktivead. Hetkel on märgendatud 1453 lauset, mida oleme kasutanud korrektuurimudelite testimisel.

Veamärgendusega korpuse arendusosa sai kasutatud tehisvigade genereerimise alusena, et tuletada erinevate vealiikide tüüpilist sagedust, mida imiteerida. Samuti oli see kasutatud reeglite väljatöötamiseks, et probleemseid vealiike tõhusamalt tuvastada ja parandada ning vähendada liigseid parandusi.

\subsection{Treeningandmed}

Masinõppe põhiste mudelite treenimiseks kasutasime TÜ keeleõppija vigade paralleelkorpust. Korpus sisaldab ligi 9000 lauset ja 128 000 sõna. Võrreldes teiste keelte vigade paralleelkorpustega on see väga väike arv: ukraina keele puhul on olemas 30 000 lauset, saksa ja tšehhi keele jaoks üle 60 000.

Lisaks inimvigu sisaldavatele andmetele kasutasime sarnaselt teiste keelte praktikaga ka tehisvigu, st korrektseid tekste kuhu on automaatselt lisatud vigu. Jõudsime keskenduda ainult lihtsate tõenäosuslike vigade sünteesimisele, mis piirduvad sõnade ja tähtede juhuslikul kustutamisel, lisamisel ja ümber paigutamisel sagedustega, mis on tuletatud arenduskorpuse inimvigade põhjal.

Projekti lõpus jõudsime teha eelkatseid ka lähenemisega, kus tehisvigade genereerimist õpetatakse närvi\-võrgu põhistele mudelitele või küsitakse promptimise kaudu suurtelt keelemudelitelt (esialgsed tulemused on leitavad 4. peatükkist).

Samuti katsetasime ka tehisvigade genereerimisel ka edasi-tagasi tõlget, mis ei jõudnud automaatse korrektuuri katseteni\footnote{Tehisvigade genereerimist edasi-tagasi tõlke meetodil kasutanud nt Lichtarge jt (2019) ja Palma Gomez jt (2023). Tõlkisime eesti keele ühendkorpusest juhuslikult valitud lauseid Neurotõlke abil soome, inglise ja vene keelde ning tagasi eesti keelde. Analüüsisime muudatusi, jaotades need vigadeks ja ümbersõnastusteks. Vene keele kaudu tõlkimine tõi kaasa rohkem vigu (74\% muudatustest) kui inglise ja soome keele kaudu tõlkimine (vastavalt 65,5\% ja 51\% muudatustest). Edasi-tagasi tõlkega saab genereerida eelkõige sõna- ja vormivaliku vigu, liigse või puuduva sõna vigu ja sõnajärjevigu. Selle meetodiga saadud treeningandmete kasulikkust eesti keele automaatkorrektuuris plaanime hinnata edaspidi.

TLÜ-s on tegemisel ka bakalaureusetöö, mis keskendub kirjavahemärgivigade genereerimisele, selleks et aidata kaasa eeskätt liigsete kirjavahemärkide kustutamisele või sobimatute kirjavahemärkide asendamisele – need on automaatkorrektuuri jaoks keerukamad vealiigid kui puuduva kirjavahemärgi lisamine.}.

\section{Metoodika}
\label{sct:meth}

\subsection{Närvivõrgupõhine grammatikakontroll}

Projekti alguse seisuga seisnes parim lähenemine grammatilisele korrektuurile jadast-jadasse teisenduste (seq2seq) masinõppimisel. Selleks kasutasime transformer-tüüpi närvivõrke. Peamiseks pudelikaelaks on seejuures andmete väike kogus, kuna transformerite hea töö jaoks on vaja kümneid miljoneid lauseid.
Projekti 1. aasta tulemused näitasid, et lihtsate tehisvigadega sünteetiliste andmete kasutamine parandab automaatkorrektuuri tulemuslikkust väikesel määral ning enamasti suurendab meetodite täpsust, kuid jätab saagise kasinaks (vt ptk 3). Samuti nägime, et alternatiivne lähenemine, mis põhineb nn ühekeelsel zero-shot masintõlkel, on oluliselt suurema saagisega, kuid väiksema täpsusega. Lahtiseks jäi küsimus, kuidas saab nende kahe tugevust kombineerida.
Projekti 2. aastal katsetasime (a) zero-shot ja tehisvigadel põhineva parandaja kombineerimist ning (b) suurte tõlkemudelite kasulikkust grammatika parandamise jaoks. Selle töö põhjal on ilmumas artikkel konverentsil EACL (Luhtaru et al., 2024).

\subsection{Keelemudelipõhine grammatikakontroll}

Kuna projekti käigus tekkis uus paradigma, nimelt suured keelemudelid ootamatute oskustega (ingl emergent abilities), tegime algseid katseid nende kasutamisega. Eesti keele tugi on projekti lõpu seisuga ainult OpenAI kommertsmudelitel GPT3.5 ja GPT4 ning seega mõõtsime nende keelemudelite promptimise efektiivsust eestikeelsete grammatiliste vigade parandamisel.

\subsection{Statistikapõhine õigekirjakontroll}

Eesti keele õigekirjakontrolli edasiarenduseks kasutasime vähem arvutuskulukat statistilist lähenemist, mis vajab treeningandmetena vaid korrektse keelekasutuse näiteid. Kõrvutasime Vabamorfi reeglipõhise spelleriga kolme keelest sõltumatut õigekirjaparanduse tööriista, mis arvestavad rohkemal või vähemal määral sõna kontekstiga: Symspelli ja Peter Norvigi bigrammidele ehk sõnapaaridele tuginevat algoritmi ning Jamspelli algoritmi, mis põhineb trigrammide ehk sõnakolmikute statistikal. Treeningmaterjali võtsime 2019. aasta ühendkorpusest: Jamspelli ja Norvigi spelleri puhul kasutasime eesti keele koondkorpuse valimit (6 mln lauset), Symspelli puhul oli vaja mahukamat andmestikku ja kasutasime kogu ühendkorpust. Parimaid tulemusi saavutasime Jamspelliga, mida katsetasime ka erineva treeningmaterjaliga: veebikorpuse tekstidega, valdavalt toimetatud tekstidega (koondkorpus, Vikipeedia korpused, teadusartiklite korpus DOAJ) ning veebi- ja kirjakeelt kombineerides (vt Allkivi-Metsoja \& Kippar, 2023).

Statistilisele õigekirjakontrollile on lisatud võimalus sõnu loendipõhiselt asendada. Ligi 2500 parandust sisaldav loend on koostatud EVKK K2 tekstides korduvate õigekirjavigade alusel. Valitud on eri tekstides vähemalt kolm korda esinevad vead. Arvestatud ei ole tekstidega, mis kuuluvad veamärgendusega test- ja arenduskorpusesse. Parandusloendit saab rakendada tööriistakomplekti iseseisva komponendina, nii saab seda teksti eeltöötluseks kasutada ka koos grammatikakontrolli mudelitega. Loendit on kasutajatel võimalik oma vajaduste jaoks kohandada ja täiendada.

\subsection{Sõnajärjepõhine vealeidja}

Lähtudes sõnaliikide n-grammide ehk järjendite sagedusest eesti keeles, saab tuvastada ebatüüpilisi sõna\-jär\-jen\-deid. Need võimaldavad sõnajärjevigade kõrval avastada ka nt vigu sõnavalikul, puuduvaid ja liigseid sõnu. Senistes katsetustes oleme aluseks võtnud sõnaliikide kolmikud koos eelneva ja järgneva kontekstiga: eelnev sõnaliik või lause algus, järgnev sõnaliik või lause lõpp. Nii tulevad eraldi esile lause algul või lõpus harvaesinevad järjendid. Sõnaliigi märgendamisel kasutame Pythoni teeki Stanza. Sõnaliigikolmikute kasutuskontekstide tõenäosuse määramiseks koostasime keelemudeli koondkorpuse põhjal. Ebatüüpiliseks lugesime kontekstid, mille tõenäosus on alla 5\%.

Praegu ei paku vealeidja võimalikke parandusi, kuid veamärgendusega arenduskorpuse ja koondkorpuse statistika põhjal on kavas välja töötada ka soovituste funktsioon. Rakenduse loomisega seoses on ilmumas kaks artiklit.

\section{Hindamine}
\label{sct:eval}

Projekti käigus valminud automaatkorrektuuri tööriistade tulemuslikkust hindasime eelkõige järgmiste kriteeriumite alusel:
\begin{itemize}
    \item veaparanduse täpsus: korrektsete paranduste osakaal tehtud paranduste hulgas;
    \item veaparanduse saagis: korrektselt parandatud vigade osakaal;
    \item F$_{0,5}$-skoor: täpsuse ja saagise kaalutud keskmine, mis annab kaks korda suurema kaalu täpsusele, eelistades parandamata jäänud vigu valeparandustele.
\end{itemize}

Korrektuurimudelite hindamiseks kohandasime eesti keelele programmi MaxMatch (M2) Scorer (Dahlmeier \& Ng, 2012). Hindamisskript võrdleb korrektori väljundit M2-formaadis ekspertparandustega ja valib iga lause puhul parandusvariandi, mille korral \fsc{}-skoor on kõige suurem. Lõpuks summeerib programm kogu testmaterjali veaparanduse täpsuse, saagise ja \fsc{}-skoori. M2-skoorija arvestab sellega, et eestikeelsetes tekstides kattuvad sõnajärjevead sageli teiste vigadega, ning hindab nende parandamist eraldiseisvalt. Näiteks kui sõnajärg on jäänud muutmata, kuid korrigeeritud on mõne vales järjestuses esineva sõna vormi, siis läheb see parandusena arvesse. Oleme rakendust täiendanud, nii et see võimaldab arvutada ka vealiikide kaupa saagise veaparanduses ja veatuvastuses (vigade osakaal, mida korrektor on üritanud parandada).

\subsection{Grammatikakontroll}

Tabel 1 sisaldab K2 testkorpusega saadud hindamistulemuste võrdlust. Kuigi suurima täpsusega on GPT4-l põhinevad parandused, on suurim \fsc{}-skoor ja saagis meie arendatud lähenemisel “NELB-1.3b”, mis kombineerib tehisandmetel eeltreenimist ja ühekeelset zero shot masintõlget (see mudel on ka lõpliku tulemina avaldatud Koodivaramus parandaja kratina). Tulemused näitavad suurt hüpet 1. ja 2. projektiaasta vahel.

Õigekirja- ja grammatikakorrektori kombineerimine üldiselt grammatikaparanduse tulemusi ei paranda, v.a ühekeelse zero-shot masintõlke puhul. Vealiikidest saab grammatikaparandaja kõige paremini hakkama just õigekirjavigade (saagis 81\%), samuti puuduva koma (saagis 79\%), käänd- ja tegusõnade vormivigade (saagis vastavalt 69\% ja 67\%) ning kokku- ja lahkukirjutuse vigade (saagis 61\%) parandamisega.

Tabel 2 näitab meie parima meetodi ning GPT4 efektiivsust erinevate keeleõppetasemete tekstide parandamisel ja emakeelena rääkijate vigade parandamisel. Meie meetod selgelt “võidab” B1- ja B2-taseme puhul, kuid “kaotab” A2- ja C1-taseme ning emakeelena rääkijate puhul. Viimane on oodatav tulemus, kuna treeningandmed sisaldavad ainult keeleõppijate vigu:

\begin{table}[t]
\centering

\begin{tabular}{|c|l|ccc|}
\hline
\textbf{Aasta} & \textbf{Mudel} & \textbf{Täpsus} & \textbf{Saagis} & \textbf{\fsc} \\ \hline
\multicolumn{5}{c}{Grammatikaparandaja} \\ \hline
2 & NELB-1.3b & 0.7192 & \bf 0.5544 & \bf 0.6788 \\ \hline
1 & eeltreenimine tehisandmetel & 0.6617 & 0.4204 & 0.5936 \\ \hline
1 & ühek. \emph{zero-shot} masintõlge & 0.5616 & 0.4361 & 0.531 \\ \hline
\multicolumn{5}{c}{Õigekirja- ja grammatikaparandaja} \\ \hline
2 & koondkorpus-vl + NELB-1.3b & 0.7148 & 0.5522 & 0.675 \\ \hline
2 & koond- ja veebikorpus-vl + NELB-1.3b & 0.7144 & 0.5521 & 0.6747 \\ \hline
2 & veebikorpus-vl + NELB-1.3b & 0.7186 & 0.5532 & 0.6781 \\ \hline
1 & veebikorpuse mudel + eeltreenimine tehisandmetel & 0.6519 & 0.4212 & 0.5875 \\ \hline
1 & koondkorpuse mudel + ühek. \emph{zero-shot} masintõlge & 0.5694 & 0.4506 & 0.5409 \\ \hline
\multicolumn{5}{c}{Võrdlus GPT4-ga} \\ \hline
2 & GPT4-prompt (zero-shot) & \bf 0.7431 & 0.4921 & 0.6743 \\  \hline
\end{tabular}
\caption{Üldised tulemused, mis saadud K2 (ehk keeleõppija vigade) testkorpusega. \emph{Aasta} viidab käesoleva projekti aastale, millal tulemused on saadud.}
\end{table}

\begin{table}[h!]
\centering
\begin{tabular}{|l|l|ccc|}
\hline
\textbf{Tase} & \textbf{Mudel} & \textbf{Täpsus} & \textbf{Saagis} & \textbf{\fsc} \\ \hline
A2 & NELB-1.3b & 0.7262 & 0.5657 & 0.6872 \\
 & GPT4-prompt & \bf 0.7546 & \bf 0.5825 & \bf 0.7125 \\ \hline
B1 & NELB-1.3b & \bf 0.7666 & \bf 0.5914 & \bf 0.7237 \\
 & GPT4-prompt & 0.7628 & 0.533 & 0.7022 \\ \hline
B2 & NELB-1.3b & \bf 0.7212 & \bf 0.5067 & \bf 0.6649 \\
 & GPT4-prompt & 0.7129 & 0.4334 & 0.6314 \\ \hline
C1 & NELB-1.3b & 0.6699 & \bf 0.5706 & 0.6473 \\
 & GPT4-prompt & \bf 0.7415 & 0.4415 & \bf 0.6528 \\ \hline
K1 & NELB-1.3b & 0.321 & 0.3904 & 0.3328 \\
 & GPT4-prompt & \bf 0.7673 & \bf 0.7190 & \bf 0.7572 \\ \hline
\end{tabular}
\caption{Tulemused keele valdamise tasemete kaupa. }
\label{tab:my_label}
\end{table}

\subsection{Õigekirjakontroll}

Spellereid oleme hinnanud kahesuguse testmaterjali alusel. 

1) Kuni valmis M2-märgendusega testkorpus, kasutasime olemasolevate õigekirjakontrolli rakenduste ja eesti keelel seni proovimata statistiliste algoritmide võrdlemiseks CoNLL-U formaadis veamärgendusega testandmeid. Need koosnevad 84 eri tasemega K2 tekstist, mis sisaldavad 1054 lauset ja 9186 sõna, sh 309 täheortograafiaveaga sõna (vt Allkivi \& Kippar, 2023). Eristasime liitvigu, mille puhul langeb õigekirjaviga kokku muud liiki veaga – selliseid oli andmestikus 46. Hindasime nii veatuvastuse kui ka veaparanduse täpsust ja saagist, võttes arvesse ka osalisi parandusi ehk õigekirja korrigeerimist, kui on eksitud nt vormi või algustähe valikul.

Esmalt kõrvutasime Vabamorfi, Jamspelli, Symspelli ja Norvigi spellerit ning Google’i ja MS Wordi korrektorit. Kui parandusvariante oli mitu, arvestasime neist esimesega. Õigekirjavigade tuvastamisel saavutasid parimaid tulemusi Vabamorf (\fsc = 84,3\%), Jamspell (\fsc = 83,9\%) ja MS Word (\fsc = 83,4\%)\footnote{Õigekirjakontrollijate testmaterjal ja väljundid asuvad lehel \url{https://github.com/tlu-dt-nlp/spell-testing}.}. Veaparanduses oli suure saagise tõttu parim Google (\fsc = 67,5\%). Järgnes Jamspell (\fsc = 64,1\%), mille täpsus oli ligilähedane (68,4\% vs. 69,2\%), kuid saagis Google’i omast väiksem (51,1\% vs. 61,2\%). Nii Jamspell kui ka Norvigi speller (\fsc = 54,1\%) ületasid veaparanduses oluliselt Vabamorfi (\fsc = 42,6\%).

Eri andmestikega treenitud Jamspelli mudelitest oli suurima saagisega nii veatuvastuses (67\%) kui ka veaparanduses (51,1\%) koondkorpuse valimiga loodud mudel. Suurima täpsusega oli 2019. aasta veebikorpusega treenitud mudel: veatuvastuses 94,3\% ja veaparanduses 74,4\%. Vabamate veebitekstide kasutamine treeningmaterjalina vähendab tarbetute paranduste hulka, ent ka leitud vigade osakaalu. Segamudelitest andis parimaid tulemusi mudel, mille treeningmaterjalis oli koond- ja veebikorpuse lausete vahekord 10:1. Selle täpsus ja saagis jäid eelmise kahe mudeli vahepeale, \fsc{}-skoor jäi veatuvastuses (82,7\%) alla koondkorpuse mudelile (83,9\%) ja veaparanduses (63,1\%) nii veebi- kui ka koondkorpuse mudelile (vastavalt 64,7\% ja 64,1\%).

Loendipõhiseid asendusi rakendades paranesid nii veatuvase kui ka -paranduse tulemused. Veatuvastuses oli \fsc{}-skoor suurim koondkorpuse mudelil ja segamudelil (ühtviisi 86,7\%), veaparanduses aga veebikorpuse mudelil (72,2\%). Koos parandusloendiga edestasid kõik kolm Jamspelli mudelit lisaks veatuvastusele ka veaparanduses Google’i korrektorit.

2) Parimaks osutunud Jamspelli mudeleid oleme hinnanud ka M2-skoorija abil mahukama K2 testmaterjaliga. Seejuures oleme arvesse võtnud üksnes täielikke parandusi ning saagist arvestanud kõigi vealiikide suhtes. Suurima veaparanduse saagise ja \fsc{}-skooriga on endiselt koondkorpusega treenitud mudel, suurima täpsusega veebikorpuse mudel. Tabelis 3 on välja toodud nende kahe mudeli parandustulemused ilma vealoendita ja koos vealoendiga (vl).

\begin{table}[h!]
\centering
\begin{tabular}{|l|ccc|}
\hline
\textbf{Jamspelli mudel} & \textbf{Täpsus} & \textbf{Saagis} & \textbf{F0.5} \\ \hline
Koondkorpus       & 0.5891 & 0.0766 & 0.2519 \\
Veebikorpus       & 0.6150 & 0.0629 & 0.2232 \\
Koondkorpus-vl    & 0.6519 & 0.0921 & 0.2942 \\
Veebikorpus-vl    & 0.6860 & 0.0867 & 0.2880 \\ \hline
\end{tabular}
\caption{Jamspelli mudeli võrdlus}
\label{table:jamspelli_model_comparison}
\end{table}

Õigekirjavigade kõrval parandab Jamspelli speller ka vormi- ja sõnavalikuvigu ning võib konteksti põhjal korrigeerida vigu, mis on olemasoleva sõna homonüümid (nt vaga ~ väga, käsime ~ käisime). Tarbetute paranduste vähendamiseks on edaspidi plaanis liidestada spelleriga nime- ja liitsõnatuvastus ning määrata reeglid, mis ei luba pakkuda algsest sõnast liialt erinevad parandust. See aitaks vältida nimede parandamist muud liiki sõnadeks (nt Juta – Juba) või sagedamini esinevateks nimedeks (nt Bosforisse – Bostonisse), samuti liitsõnaosade ebavajalikku asendamist (nt digioskusi – digiboksi).

\subsection{Sõnajärjepõhine veatuvastus}

Sõnajärjepõhise vealeidja hindamiseks leidsime veaohtliku sõnajärjendi asukoha lauses ning võrdlesime seda M2-märgendusega. Kui samas piirkonnas oli veamärgenduse alusel viga, siis lugesime veatuvastuse õigeks. Juhul kui sõnaliigikolmik esines kontekstis, mille tõenäosus on keelemudeli järgi vähem kui 5\%, oli veatuvastuse täpsus 75\%, saagis 39,2\% ja \fsc{}-skoor 63,4\%. Veaohtlikkuse tõenäosuspiiri vähendades suureneb täpsus, aga kahaneb saagis. Rakendus võimaldab sageli avastada eksimusi V2-sõnajärje vastu lause või osalause algul, olema-tegusõna puudumist ja sõnakordusi, ent ka muid sõnajärje-, puuduva ja liigse sõna vigu.

Proovisime vigu tuvastada ka kõige parema grammatikakontrolli mudeli (NELB) väljundist. Vealeidja tõi esile 2219 probleemset sõnajärjendit 1031 lauses. Seni ei ole me täpsemini hinnanud, millal on veatuvastus õigustatud. Selleks et tuvastustäpsust parandada, on kavas kindlaks teha sõnajärjendid ja kontekstid, millega väiksest tõenäosusest hoolimata tavaliselt viga ei kaasne. Näiteks on asesõnu sisaldavad järjendid keeleõppijate tekstides sagedamad kui eesti kirjakeeles üldiselt, ent tegemist pole vigadega.

Samas tasub kõigi valminud korrektuuritööriistade puhul kvalitatiivselt vaadelda juhtumeid, kus korrektor on lauset muutnud või tuvastanud vea, aga asendus ei kattu veamärgendusega või polegi viga märgendatud. See töö on plaanis jätkuprojektis.

\section{Tulevikusuunad}
\label{sct:fut}

\subsection{Andmete vajadused}

Meie projekti tulemused näitavad, et treenimisandmeid on siiski vaja rohkem, eriti A2- ja C1-taseme keeleõppijate vigade ning emakeelena rääkijate vigadega. Orienteeruvat suurust võib võtta tšehhi keelelt, mille põhjal usume, et 50 000 – 60 000 lause lisamine võiks oluliselt suurendada nende vigade parandamise efektiivsust.

Andmeid saab ka genereerida suurte keelemudelite abil. Idee seisneb promptimise teel saadud tulemustes, kus promptis on antud korrektne lause ning keelemudeli ülesandeks on lisada sinna vigu, mida võiks teha nt B1-taseme keeleõppija või emakeelena rääkija.

Tegime esialgseid katseid tehisvigade genereerimise õpetamisega tehisnärvivõrkudele, kus võtsime K2 tekstide ja vigadega treenimiskorpust (mida terve projekti käigus kasutasime) ning asendasime inimvigu GPT3.5 ja GPT4 abil genereeritud vigadega. Seejärel treenisime väikse korrektuurimudeli kõigi kolme treeningkorpusega (inimvead, GPT3.5 vead ja GPT4 vead: kõigi kolme puhul 9’000 lauset). Võrdluseks on ka mudel, mis on õppinud 1 miljoni lausega, mis sisaldavad ainult tõenäosuslike vigu. Tabelist 4 on näha, et GPT mudelitega saadud tehisvigade abil jõuab peaaegu sama kaugele kui inimvigadega; tõenäosuslikud vead jäävad aga saagisega kaugele taha, vaatamata suuremale materjalihulgale:

\begin{table}[ht]
\centering
\begin{tabular}{|l|ccc|}
\hline
\textbf{Vigade allikas} & \textbf{Täpsus} & \textbf{Saagis} & \textbf{F0.5} \\ \hline
Inimvead (9k lauset) & 0.5849 & 0.4133 & \bf 0.5401 \\  \hline
GPT3.5 vead (9k lauset) & 0.5859 & 0.3725 & 0.5257 \\
GPT4 vead (9k lauset) & 0.5525 &  \bf 0.4309 & 0.5230 \\  \hline
Tõenäosuslikud vead (1M lauset) &  \bf 0.6866 & 0.1414 & 0.3876 \\ \hline
\end{tabular}
\caption{Erinevate vigade allikate mõju korrektuuri tulemustele, väikese korrektuurimudeliga.}
\label{tab:errors_analysis}
\end{table}

Selle põhjal usume,  et suurte keelemudelite genereeritud tehisvead peaksid olema palju kasulikumad kui statistilised tehisvead ning sel viisil on võimalik lisada mitu sada tuhat lauset koos vigade ja parandustega.

\subsection{Automaatkorrektuur suurte keelemudelitega}

Nagu näitavad meie hindamistulemused, on GPT4 võimeline parandama eestikeelseid vigu heal tasemel. Selle puudujäägid on:

\begin{itemize}
    \item tegu on kommertstootega;
    \item selle kasutamine kaotab privaatsust;
    \item meie näitasime, et isegi ilma keelemudeliteta on võimalik saavutada kõrgemat parandamise korrektsust.
\end{itemize}

Eeldades, et lähiajal tekib ka vabavaralisi keelemudeleid eesti keele toega, annab meie arust kõige suuremat lootust keelemudelite kasutamise puhul nende peenhäälestamine eksplitsiitselt vigade parandamise ülesandele.

\subsection{Automaatkorrektuuri mudelite efektiivsus}

Keelemudelid on äärmiselt ebaefektiivsed: nad on palju suuremad ja aeglasemad kui ülesande-spetsiifilised lahendused. Keelemudeleid aga saab kasutada andmete loomiseks mudelite treenimise ajal, nt parandusandmete genereerimiseks (kirjeldatud üleval).

Juhul kui on vaja parandajat, mis nõuab vähe mälu või suudab töötada pisiseadmete sees, siis tasub arendada ka distilleeritud jadast-jadasse teisendusi teostatavaid närvivõrke ning samuti ka statistilisi meetodeid.

\subsection{Lisafunktsionaalsuse arendamine}

Suurte keelemudelite ootamatud oskused võimaldavad arendada lisafunktsionaalsust. Käesolevas projektis jäi katmata vealiikide automaatne tuvastamine; selliste andmete käsitsi märgendamise asemel on võimalik neid andmeid genereerida suure keelemudeli abil (nt GPT4) ning seejärel treenida liigitaja nende andmete põhjal.

Keeleõppimise abi jaoks on vaja pakutud grammatiliste paranduste selgituste genereerimine. Selle ülesan\-dega saavad hakkama keelemudelid ise. Tasub uurida ka selle ülesande distilleerimise võimalust.

\section*{Projekti artiklid}

\hspace{4.5mm} Luhtaru, A., Korotkova, E. \& Fishel, M. (2024). No Error Left Behind: Multilingual Grammatical Error Correction with Pre-trained Translation Models. Accepted to: Proceedings of EACL’2024: The 18th Conference of the European Chapter of the Association for Computational Linguistics. ACL Anthology, in print. \url{https://openreview.net/forum?id=vchiWnuieL}

Allkivi-Metsoja, K. \& Kippar, J. (2023). Spelling Correction for Estonian Learner Language. Proceedings of the 24th Nordic Conference on Computational Linguistics. ACL Anthology, 782–788. \url{https://aclanthology.org/2023.nodalida-1.79.pdf}

\section*{Teised viited}

\hspace{4.5mm} Bryant, C., Felice, M., Andersen, O. E., \& Briscoe, T. (2019). The BEA-2019 Shared Task on Grammatical Error Correction. Proceedings of the Fourteenth Workshop on Innovative Use of NLP for Building Educational Applications. ACL Anthology, 52–75. \url{http://dx.doi.org/10.18653/v1/W19-4406}

Bryant, C., Felice, M., \& Briscoe, T. (2017). Automatic Annotation and Evaluation of Error Types for Grammatical Error Correction. Proceedings of the 55th Annual Meeting of the Association for Computational Linguistics, Vancouver, Canada. ACL Anthology, 793–805. \url{https://doi.org/10.18653/v1/P17-1074}

Dahlmeier, D., \& Ng, H. T. (2012). Better Evaluation for Grammatical Error Correction. Proceedings of the Conference of the North American Chapter of the Association for Computational Linguistics: Human Language Technologies, Montréal, Canada. ACL Anthology, 568–572. \url{https://aclanthology.org/N12-1067.pdf}

Lichtarge, J., Alberti, C., Kumar, S., Shazeer, N., Parmar, N., \& Tong, S. (2019). Corpora Generation for Grammatical Error Correction. Proceedings of NAACL-HLT 2019, Minneapolis, Minnesota, June 2 – June 7, 2019. ACL Anthology, 329–3301. \url{http://dx.doi.org/10.18653/v1/N19-1333}

Palma Gomez, F., Rozovskaya, A., \& Roth, D. (2023). A Low-Resource Approach to the Grammatical Error Correction of Ukrainian. Proceedings of the Second Ukrainian Natural Language Processing Workshop (UNLP). ACL Anthology, 114–120. \url{https://aclanthology.org/2023.unlp-1.14.pdf}

\end{document}